\documentclass[aps,prb,twocolumn,amsmath,amssymb,superscriptaddress,floatfix]{revtex4-1}

\usepackage{graphicx}
\usepackage{bm}
\usepackage{hyperref}
\usepackage{braket,bbold,color}
\usepackage{dcolumn}
\usepackage{tikz} 
\usepackage{ifthen}
\usetikzlibrary{calc}

\begin{document}

\title{Algorithms for Tensor Network Contraction Ordering}

\author{Frank Schindler}
\affiliation{Department of Physics, University of Zurich, Winterthurerstrasse 190, 8057 Zurich, Switzerland}
\affiliation{Kavli Institute for Theoretical Physics, University of California, Santa Barbara, CA 93106, USA}

\author{Adam S. Jermyn}
\email{adamjermyn@gmail.com}
\affiliation{Center for Computational Astrophysics, Flatiron Institute, New York, NY 10010, USA}
\affiliation{Kavli Institute for Theoretical Physics, University of California, Santa Barbara, CA 93106, USA}

\date{\today} 

\begin{abstract}
Contracting tensor networks is often computationally demanding. Well-designed contraction sequences can dramatically reduce the contraction cost. We explore the performance of simulated annealing and genetic algorithms, two common discrete optimization techniques, to this ordering problem. We benchmark their performance as well as that of the commonly-used greedy search on physically relevant tensor networks. Where computationally feasible, we also compare them with the optimal contraction sequence obtained by an exhaustive search. We find that the algorithms we consider consistently outperform a greedy search given equal computational resources, with an advantage that scales with tensor network size. We compare the obtained contraction sequences and identify signs of highly non-local optimization, with the more sophisticated algorithms sacrificing run-time early in the contraction for better overall performance.
\end{abstract}


\maketitle

\section{Introduction}
Tensor networks are a convenient language for studying the statistics of discrete systems with local interactions. The partition function and correlation functions of many lattice models may be written as tensor networks. Similarly, typical states of quantum systems often admit an efficient representation as a tensor network, either in the form of matrix product states (MPS)~\cite{PhysRevLett.69.2863,ORUS2014117} or more general states such as tree tensor networks~\cite{doi:10.1063/1.4798639,XuTrees19} and projected entangled pair states (PEPS)~\cite{doi:10.1146/annurev-conmatphys-020911-125018}. Tensor networks have also been used as machine learning classifiers~\cite{NIPS2016_6211,Liu_2019}.

At the core of these applications is the problem of tensor network contraction, in which all intermediate bonds in a tensor network are summed to evaluate the network. Because these sums are performed simultaneously, a naive tensor network contraction uses computational resources which are exponential in system size, and so better approaches are needed.

Unfortunately, the problem of contracting tensor networks lies in the computational complexity class NP~\cite{doi:10.1142/S0129626497000176} and so it is likely that in the worst case exponential resources will always be needed. Nonetheless, it is often possible to approximate tensor network contraction, resulting in efficiently-computable answers with controllable errors~\cite{variationalPfeifer14,2017arXiv170903080J,doi:10.1146/annurev-conmatphys-020911-125018,ran2017lecture,2019arXiv191203014P,ClementTRG18,MoritaTRG18}. Moreover, many useful tensor networks, including MPS networks~\cite{PhysRevLett.69.2863}, can be contracted exactly in polynomial time by taking advantage of the property that the computational cost of contracting a tensor network depends strongly on the order of summation while the result does not. Hence while the worst cases may be intractable, there is still room to improve in typical or special cases.

To that end, we examine two algorithms which are widely used in discrete optimization. Our aim is to see if these algorithms provide any improvement over standard methods and hand-crafted contraction sequences. These are Genetic Algorithms~\cite{SADEGHI2014126,mitchell1996an} and Simulated Annealing~\cite{1981AcCrA..37..742K,Bollweg1997}. We begin in Section~\ref{sec:tn} by reviewing the structure of tensor networks, Penrose notation, and the computational cost of contracting sequences. In Section~\ref{sec:algs} we then describe the algorithms in more detail, along with the commonly-used Greedy Search algorithm~\cite{GASmith2018} and the reference Exhaustive Search method~\cite{Pfeifer14,NconPfeifer14,SchuylerContraction18,DumitrescuTreeExact18}. We then perform numerical experiments in Section~\ref{sec:exp}, testing these methods on both two-dimensional square tensor networks and Erd\H{o}s-R\'{e}nyi random networks, and find that both algorithms outperform the Greedy Search in most of our experiments, often by many orders of magnitude. We examine specific contraction sequences in Section~\ref{sec:seq} to understand how these algorithms craft such efficient sequences, and conclude with a discussion of our results in Section~\ref{sec:conc}.

\section{Tensor Network Contraction}
\label{sec:tn}
A tensor network is a list of tensors along with a specification of which pairs of their indices are meant to be contracted. So for instance,
\begin{align}
	N_{klmn} = \sum_{ij} T_{ijkl} X_{i} Y_{jmn}
	\label{eq:first}
\end{align}
specifies a network $N$ formed of three tensors $T$, $X$ and $Y$, with two contracted pairs of indices, namely $i$ and $j$. The network itself is tensor-valued, with the four indices $k,l,m$ and $n$, which correspond to the indices of the constituent tensors which were not contracted.

A key feature of tensor network contraction is that individual summations commute. That is, the sums in equation~\eqref{eq:first} may be done simultaneously, but we could also perform first the sum over $i$, producing the intermediate tensor
\begin{align}
Q_{jkl} = \sum_i T_{ijkl} X_i,
\end{align}
and only then perform the sum over $j$ to evaluate
\begin{align}
	N_{klmn} = \sum_j Q_{jkl} Y_{jmn}.
	\label{eq:second}
\end{align}
Likewise, we could first sum over $j$, producing
\begin{align}
Q'_{iklmn} = \sum_j T_{ijkl} Y_{jmn},
\end{align}
and then sum over $i$ to obtain
\begin{align}
	N_{klmn} = \sum_i Q'_{iklmn} X_{i}.
\end{align}
Both pathways arrive at the same answer, but they may have very different computational costs. For instance, the intermediate $Q'$ has a higher rank (number of indices) than the intermediate $Q$, and so if all bonds have the same dimension, $Q'$ requires more memory to store and more computation time to evaluate.

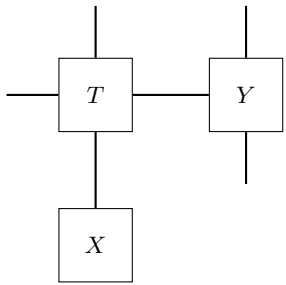
\begin{figure}\centering
	\begin{tikzpicture}

		\node[draw=none] (v0) at (-1,1.3) {};
		\node[draw=none] (v1) at (1,1.3) {};
		\node[draw=none] (v2) at (-1,-1.3) {};
		\node[draw=none] (v3) at (1,-1.3) {};
		\node[draw=none] (t0) at (-2.3,0) {};
		
	    \node[rectangle,minimum width = 3em, minimum height = 3em, draw] (u0) at (-1,0) {$T$};
	    \node[rectangle,minimum width = 3em, minimum height = 3em, draw] (u1) at (1,0) {$Y$};
	    \node[rectangle,minimum width = 3em, minimum height = 3em, draw] (u2) at (-1,-2) {$X$};

		\draw[thick]
			(u0) -- (u0 -| u1.west)
			(t0) -- (t0 -| u0.west)
			(v0) -- (v0 |- u0.north)
			(v1) -- (v1 |- u1.north)
			(u2) -- (u2 |- u0.south)
			(v3) -- (v3 |- u1.south);

	\end{tikzpicture}
	\caption{The tensor network specified by equation~\eqref{eq:first} in Penrose notation.}
	\label{fig:net1}
\end{figure}

It is often convenient to write small tensor networks explicitly, as in equation~\eqref{eq:first}, but for large ones this quickly becomes cumbersome. Instead we depict larger networks graphically using Penrose notation, with squares representing tensors and lines representing indices~\cite{Penrose}. So, for instance, the network specified by the right-hand side of equation~\eqref{eq:first} is shown graphically in Figure~\ref{fig:net1}.

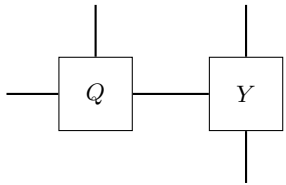
\begin{figure}\centering
	\begin{tikzpicture}

		\node[draw=none] (v0) at (-1,1.3) {};
		\node[draw=none] (v1) at (1,1.3) {};
		\node[draw=none] (v2) at (-1,-1.3) {};
		\node[draw=none] (v3) at (1,-1.3) {};
		\node[draw=none] (t0) at (-2.3,0) {};
		
	    \node[rectangle,minimum width = 3em, minimum height = 3em, draw] (u0) at (-1,0) {$Q$};
	    \node[rectangle,minimum width = 3em, minimum height = 3em, draw] (u1) at (1,0) {$Y$};

		\draw[thick]
			(u0) -- (u0 -| u1.west)
			(t0) -- (t0 -| u0.west)
			(v0) -- (v0 |- u0.north)
			(v1) -- (v1 |- u1.north)
			(v3) -- (v3 |- u1.south);

	\end{tikzpicture}
	\caption{The tensor network specified by equation~\eqref{eq:second} in Penrose notation.}
	\label{fig:net2}
\end{figure}

\begin{figure}\centering
	\begin{tikzpicture}

		\node[draw=none] (v0) at (-1,1.3) {};
		\node[draw=none] (t0) at (-2.3,0) {};
		\node[draw=none] (v1) at (-1,-1.3) {};
		\node[draw=none] (t1) at (0.3,0) {};
		
	    \node[rectangle,minimum width = 3em, minimum height = 3em, draw] (u0) at (-1,0) {$N$};

		\draw[thick]
			(t0) -- (t0 -| u0.west)
			(v0) -- (v0 |- u0.north)
			(t1) -- (t1 -| u0.east)
			(v1) -- (v1 |- u0.south);

	\end{tikzpicture}
	\caption{The tensor $N$ appearing in equation~\eqref{eq:first} in Penrose notation.}
	\label{fig:net3}
\end{figure}
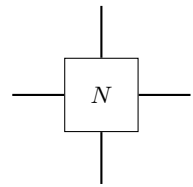

In this notation, performing a single sum amounts to combining two nodes in the graph into one. Hence, summing over $j$ in equation~\eqref{eq:first} produces the network shown in Figure~\ref{fig:net2}. Then, summing over $i$ finally yields the evaluated network shown in Figure~\ref{fig:net3}.

We call the order in which we contract pairs of indices a contraction sequence. To calculate the computational cost of a given contraction sequence, we count the number of floating-point multiplications that have to be performed~\cite{Pfeifer14}. This is equal to the number of floating-point additions, and so counts the number of operations required and the run-time up to a constant factor. Our cost function thus reads
\begin{equation}
\mathrm{cost}(\{E\}) = \sum_{e \in \{E\}} \prod_{m \in \{v_e\}} \chi(m),
\label{eq:cost}
\end{equation}
where $\{E\}$ denotes the ordered set of edges to be contracted, $\{v_e\}$ denotes the set of edges adjoining the two vertices connected by the edge $e$ (including $e$ itself) at a given contraction step, and $\chi(m)$ is the bond dimension of edge $m$, i.e., the number of different values that the index associated with $m$ can assume. Holding coordination number fixed, the computational complexity of evaluating this cost function is $\mathcal{O}(E)$ where $E$ is the number of edges in the network.

As an example, consider the contraction in Eq.~\eqref{eq:first}. Contracting the index $i$ first amounts to $\chi(i)\chi(j)\chi(k)\chi(l)$ elementary operations. Following up with the sum over the index $j$ then adds another $\chi(j)\chi(k)\chi(l)\chi(m)\chi(n)$ operations. The total cost of this contraction ordering is therefore 
\begin{align}
\rm{cost}_{ij}=\chi(i)\chi(j)\chi(k)\chi(l) + \chi(j)\chi(k)\chi(l)\chi(m)\chi(n).
\end{align}
By contrast, the cost of contracting first $j$ then $i$ is
\begin{align}
\rm{cost}_{ji} = &\ \chi(i)\chi(j)\chi(k)\chi(l)\chi(m)\chi(n) \nonumber\\&+ \chi(i)\chi(k)\chi(l)\chi(m)\chi(n).
\end{align}
These are clearly different, and so there is a room to optimize by picking the lower-cost option.

\section{Algorithms}
\label{sec:algs}
We have tested four algorithms. Two of these, namely Exhaustive Search and Greedy Search, are in common use for obtaining tensor network contraction sequences. So far as we are aware, the other two have not previously been used to this end.

The other algorithms, namely Simulated Annealing and the Genetic Algorithm, share a common structure which will aid in their comparison. Each consists of a procedure for generating batches of contraction sequences. Only one batch is considered at a time. The cost of contracting the given tensor network with each sequence in the batch is then computed. If any sequence in the proposal has a lower cost than the previous lowest-cost sequence it is stored in place of the previous best sequence. The algorithm then proceeds to propose a new batch, possibly using the results of the previous batches. This process iterates until either all possible orderings have been considered or a time limit is reached. Where these methods differ is in the rule for producing new batches.

\begin{figure}[t]
\centering
\includegraphics[width=\columnwidth,page=4]{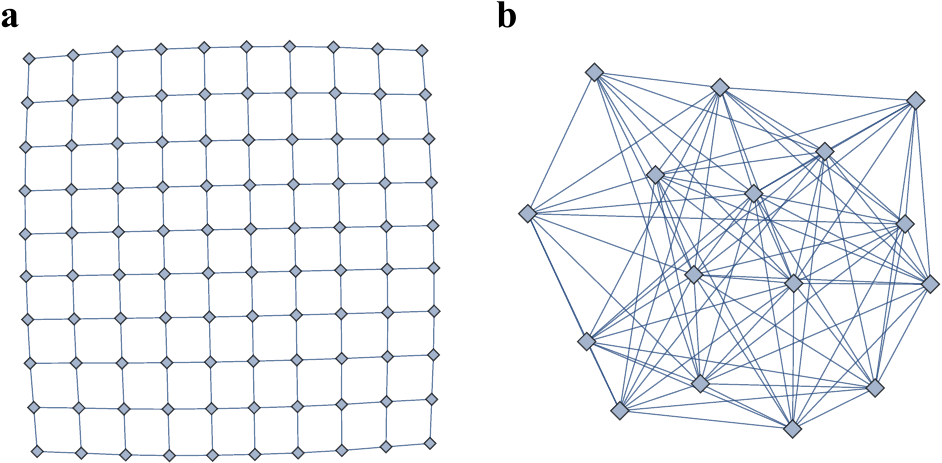}
\caption{Examples of the test tensor networks used to compare the different algorithms in Penrose notation. (a)~Square tensor network with $10^2$ nodes. (b)~Random tensor network with $16$ nodes.}
\label{fig:testnetworks}
\end{figure}

Because of this structure both methods may be run for as long as desired and can at any point return the best ordering found so far. We will take advantage of this to restrict each method to a fixed number of evaluations of the cost function. This limitation is a proxy for a runtime limit that is insensitive to details of the implementation of the algorithm or the underlying hardware, which makes it a useful means of comparison.

We now detail the four algorithms. Implementations for the Greedy Search, the Genetic Algorithm, and Simulated Annealing can be found at \href{https://github.com/frankschindler/OptimizedTensorContraction}{github.com/frankschindler/OptimizedTensorContraction}.

\subsection{Exhaustive Search}
Exhaustive Search comes in several varieties. In its most basic version every possible contraction ordering is considered exactly once. The cost of each is evaluated and the ordering with the lowest cost is returned.

This algorithm is deterministic and always returns the optimal contraction sequence. Because the number of contraction sequences to consider scales like $\mathcal{O}(e^{E})$, where again $E$ is the number of edges in the network, the run-time of this algorithm is exponential. More advanced variants of this algorithm incorporate tree pruning~\cite{Pfeifer14} to avoid considering sequences which can be proven to have higher cost than others, but in the worst case the cost is still exponential.

For the numerical results we present for the Exhaustive Search, we adapted the MATLAB version of the Netcon algorithm from Ref.~\onlinecite{Pfeifer14} to also output the accumulated number of cost function evaluations. We then ran it with the parameter choice $\mathsf{costType}=1$, $\mathsf{muCap}=1$, $\mathsf{allowOPs}=\mathsf{false}$. We used the MATLAB version R2019a and Netcon version 2.01.

\subsection{Greedy Search}
Greedy Search begins by considering the cost of performing just one step in the contraction. Evaluating this incremental cost takes time which is $\mathcal{O}(1)$. Each possible first step is considered, and the one with the lowest cost is taken. The method proceeds recursively, considering next all possible second steps.

Alternate variants of Greedy Search have been used which consider multiple steps simultaneously~\cite{Pfeifer14,GASmith2018}. For instance one could consider all possibilities for the next two steps, or more generally for the next $k$ steps. The cost of this algorithm considering $k$ steps simultaneously is $\mathcal{O}(E^{k})$ incremental cost function evaluations, which is equivalent to $\mathcal{O}(E^{k-1})$ evaluations of the full cost function. Because the cost grows rapidly with $k$ we only consider the commonly-used~\citep{GASmith2018} case of $k=2$ in the following.

For the numerical results we present for the Greedy Search and the remaining algorithms, we made use of Python 3.7.4, with the libraries \emph{Numpy} 1.17.2, \emph{Scipy} 1.3.1, \emph{itertools}, and \emph{copy}. We implemented the $k$-step Greedy Search as a standalone Python function built on these tools.

\subsection{Genetic Algorithm}
The Genetic Algorithm begins by evaluating the fitness (the negative cost) of each contraction in a starting population of randomly generated sequences~\cite{SADEGHI2014126,mitchell1996an}. It then samples a new population, drawing from the starting population with replacement and with probabilities that are proportional to the fitness of the individual contraction sequences. This models the extinction of unfit specimen. Furthermore, the contractions in the new population are subject to mutation, in that there is a finite chance that the ordering of two randomly selected edges is exchanged in the respective sequence (the fittest individual of the population is always kept unchanged). This process is then iterated.

We implemented the Genetic Algorithm in Python. We chose a population size of $20$ and a mutation rate of $60\%$. For the fitness function, we used
\begin{align}
&\mathrm{fitness}(\{E\}) \nonumber \\&= \exp{\left[\frac{\log\mathrm{cost}(\{E_\mathrm{max}\})-\log\mathrm{cost}(\{E\})}{\log\mathrm{cost}(\{E_\mathrm{max}\})-\log\mathrm{cost}(\{E_\mathrm{min}\})}\right]}-0.99,
\end{align}
where $\{E_\mathrm{min}\}$ and $\{E_\mathrm{max}\}$ are the contraction sequences with the lowest and highest cost in the population, respectively. This fitness function was chosen heuristically to return natural values in the range $(0,e-1)$ while retaining the hierarchy of scales resolved by the original cost function (up to a power). Note that the subtraction of $0.99$ matters because we generate probabilities from a population's fitness distribution after normalization. We checked that the performance of the Genetic Algorithm is not sensitive to the precise choice of fitness function.

\subsection{Simulated Annealing}
Simulated Annealing works with an alternative representation of contraction sequences, where we encode permutations of edge labels by arrays of real numbers taken from the interval $[0,1]$. A contraction can then be obtained from the permutation that orders the numbers in the respective array by magnitude. This representation has the advantage that it allows for a continuous deformation of the arrays while the constraint that each element represent a valid permutation is implicitly taken into account. This allows us to use the dual annealing~\cite{Xiang97} variant, which combines the standard classical annealing algorithm with a local optimizing search.

For the numerical results we present for Simulated Annealing, we employed the implementation of the dual\_annealing algorithm that is available from the $\emph{optimize}$ package of the \emph{Scipy} library. We used the default settings of the algorithm, which are $\mathsf{local\_search\_options}=\{\}$, $\mathsf{initial\_temp}=5230.0$, $\mathsf{restart\_temp\_ratio}=2*10^{-5}$, $\mathsf{visit}=2.62$, $\mathsf{accept}=-5.0$, $\mathsf{seed}=\mathsf{None}$, $\mathsf{no\_local\_search}=\mathsf{False}$, $\mathsf{callback}=\mathsf{None}$, $\mathsf{x0}=\mathsf{None}$.

\section{Numerical Experiments}
\label{sec:exp}
We perform our numerical experiments on two classes of tensor networks. The first are two-dimensional square lattices, shown in Figure~\ref{fig:testnetworks}~(a). We choose two-dimensional networks because in one-dimension the optimal contraction sequence is already known, so this provides one of the simplest non-trivial test cases.

The second class of network we consider is that of Erd\H{o}s-R\'{e}nyi random graphs. These consist of a collection of nodes with edges distributed amongst them at random, such that all pairs of nodes have the same probability of having an edge, and such that edges are placed independently of one another. In our tests we let this probability be $80\%$. An example of a tensor network generated in this way is shown in Figure~\ref{fig:testnetworks}~(b). Such networks are analogous to spin glasses, and represent some of the most difficult tensor networks to contract due to their large connectivity and high variance in tensor rank.

\begin{figure}[t]
\centering
\includegraphics[width=\columnwidth,page=1]{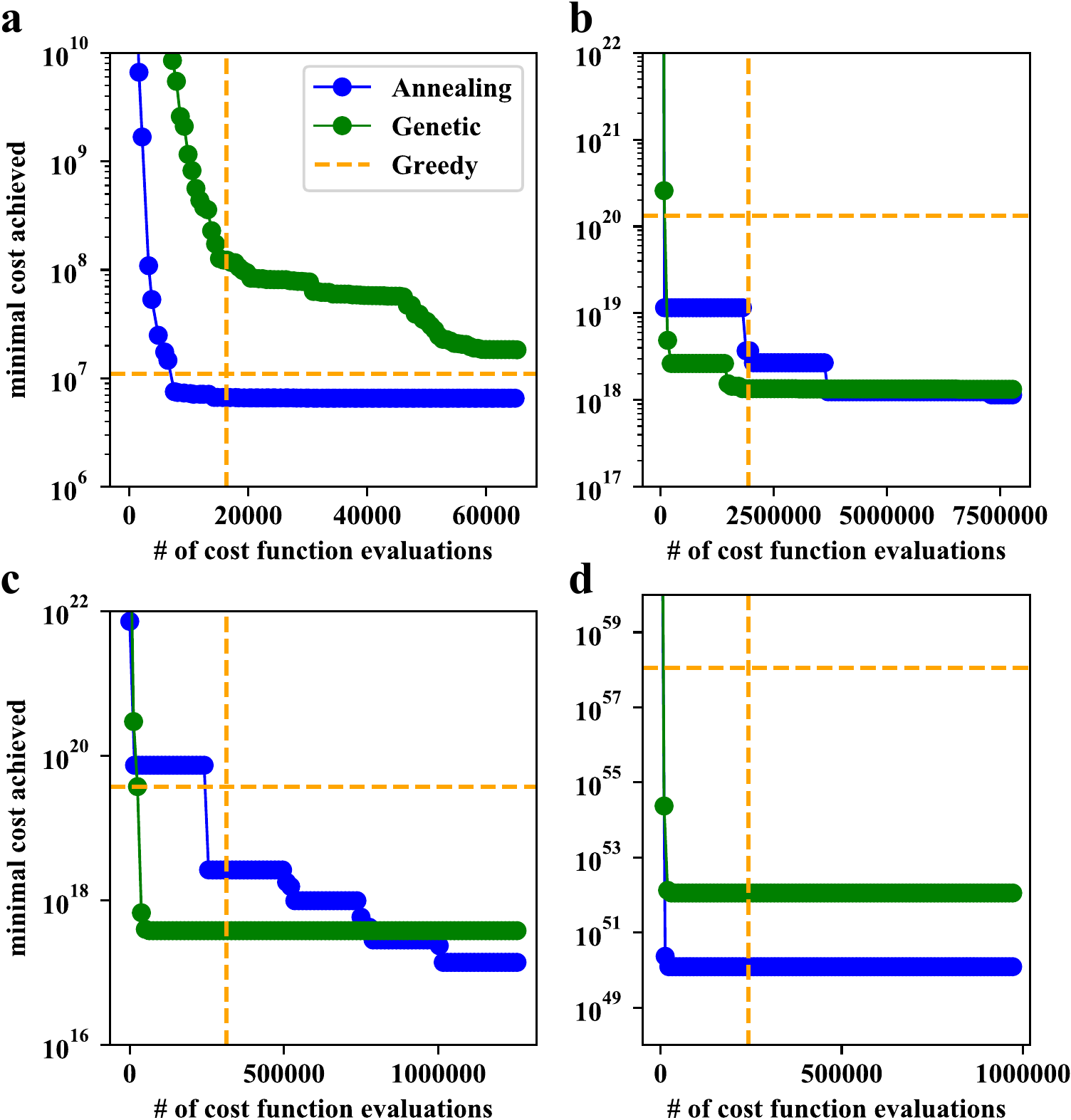}
\caption{Performance comparison by number of cost function evaluations, a platform-independent measure of algorithm runtime. Results for (a)~a square tensor network with $10^2$ nodes connected by edges with $\chi = 2$, (b)~a square tensor network with $10^2$ nodes connected by edges with $\chi = 10$, (c)~a random tensor network with $16$ nodes, where each edge (with $\chi = 2$) had a 80\% chance of being realized, and (d)~a random tensor network with $16$ nodes, where each edge (with $\chi = 10$) had a 80\% chance of being realized.}
\label{fig:performancebyruntime}
\end{figure}

\begin{figure*}[ht]
\centering
\includegraphics[width=\textwidth,page=2]{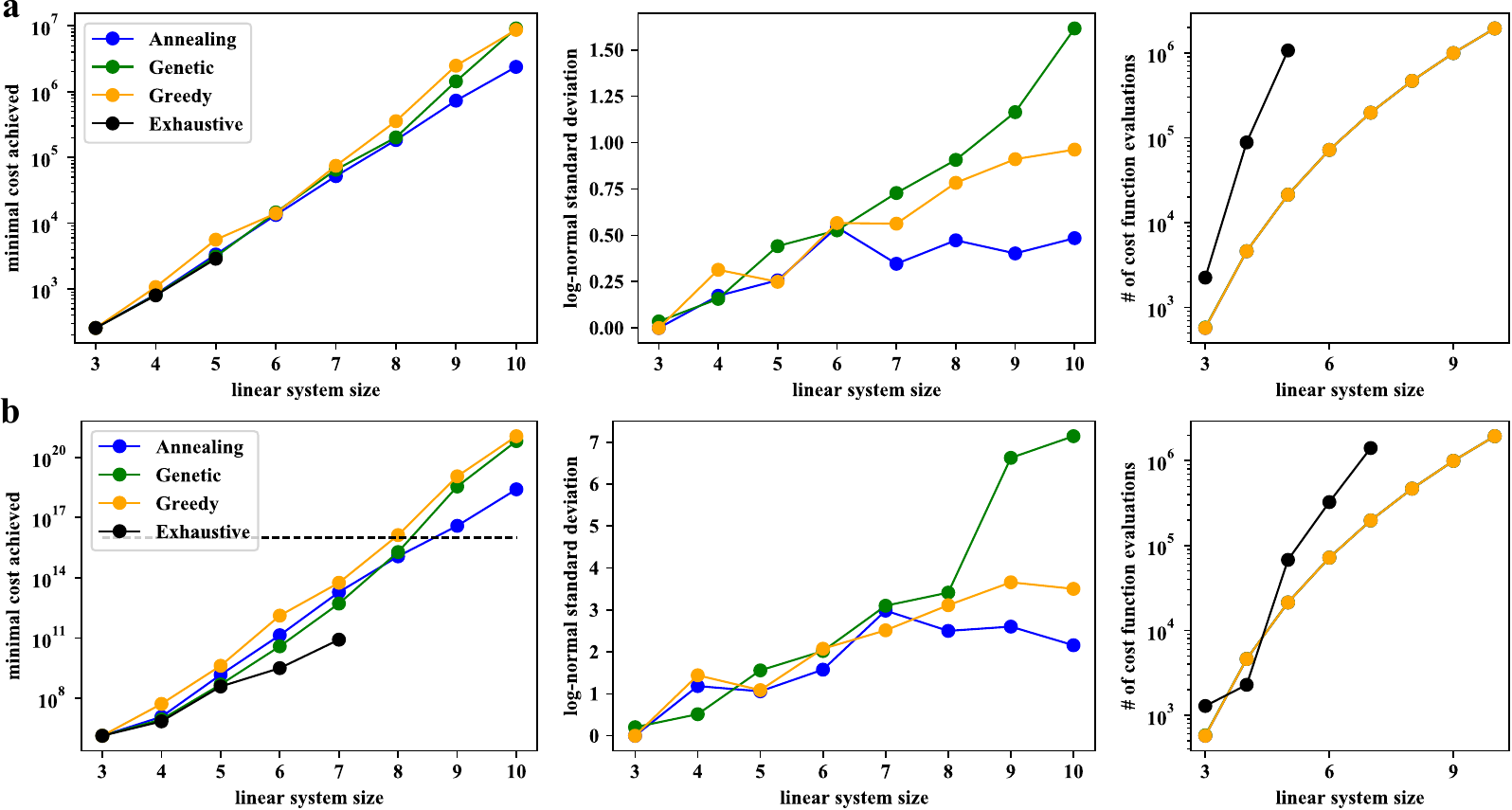}
\caption{Performance comparison for a contraction of square tensor networks of different size. (a)~$\chi = 2$. (b)~$\chi = 10$. ``Linear system size" denotes here the square root of the total number of nodes in the tensor network. The black horizontal line in the lower left panel indicates the ``desktop limit'' (see main text) of about $10^{16}$  addition and multiplication operations. For each algorithm, we show the median run out of $20$ runs, with the log-normal standard deviation (relative standard deviation) indicated in the center panels. The right panels show the number of cost function evaluations, where Simulated Annealing and the Genetic Algorithms were allowed the same number of evaluations as used by the Greedy search.}
\label{fig:performancesquare}
\end{figure*}

\begin{figure*}[t]
\centering
\includegraphics[width=\textwidth,page=3]{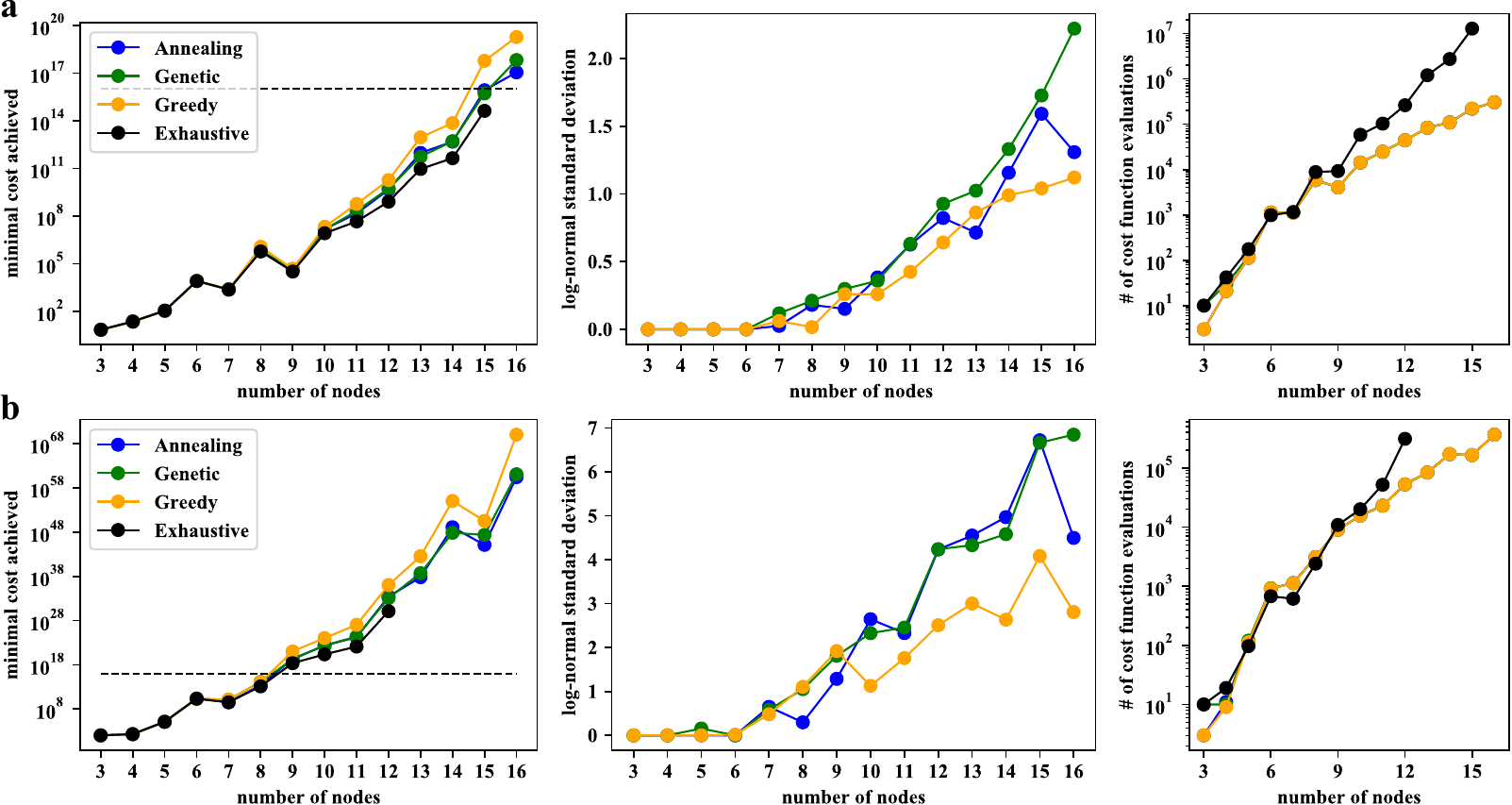}
\caption{Performance comparison for the contraction of Erd\H{o}s-R\'{e}nyi random tensor networks of different size, where, for each network, every possible edge had a 80\% chance of being realized. (a)~$\chi = 2$. (b)~$\chi = 10$. The black horizontal line in the lower left panel indicates the ``desktop limit'' (see text) of about $10^{16}$  addition and multiplication operations. For each algorithm, we show the median run out of $20$ runs, with the log-normal standard deviation (relative standard deviation) indicated in the center panels. The right panels show the number of cost function evaluations, where Simulated Annealing and the Genetic Algorithm were allowed the same number of evaluations as used by the Greedy search.
}
\label{fig:performancerandom}
\end{figure*}

\subsection{Variable Run-Time}
In our first experiment we consider the square network shown in Figure~\ref{fig:testnetworks}~(a), with a bond dimension of $\chi=2$. We use each of Simulated Annealing, the Genetic Algorithm, and Greedy Search to produce contraction sequences for this network. The results are shown in Figure~\ref{fig:performancebyruntime}~(a). For Simulated Annealing and the Genetic Algorithm we show the contraction cost given by equation~\eqref{eq:cost} of the best contraction sequence found as a function of the number of cost function evaluations used. The Greedy Search requires a fixed number of evaluations, and so we just show its output with that number of evaluations.

From this experiment we see that Simulated Annealing significantly outperforms the Genetic Algorithm when given the same number of function evaluations. This is not universally true, but we see the same in almost every case. We also see that the Greedy Search performs better than the Genetic Algorithm, but worse than Simulated Annealing at the same number of function evaluations.

Figure~\ref{fig:testnetworks}~(b) shows the same experiment but with an increased bond dimension of $\chi=10$. Increasing the bond dimension dramatically raises the contraction cost for each algorithm. The change in cost is of order $(\chi_{\rm new}/\chi_{\rm old})^{2L}$, where $L$ is the linear size of the network. This may be seen by noting that each tensor at an intermediate stage represents a contiguous subset of the original network. Eventually those subsets come to be extensive in size and so come to have perimeter length of order $L$. Hence at some point each algorithm must contract two tensors with of order $L$ bonds, with cost of order $\chi^{2L}$.

Interestingly, with larger bond dimension Greedy Search performs worse relative to the other algorithms. We understand this as follows: for small bond dimensions it is possible for many contraction steps to matter in the total cost, because the difference between contractions of different ranks is small. As the bond dimension increases the cost of a contraction sequence comes to be dominated by the cost of the few most expensive contraction step(s). Optimizing a contraction sequence then becomes mostly a matter of avoiding the worst cases. Because the appearance of very expensive contraction steps is a function of the entire contraction sequence up to that point, this is a non-local optimization problem that Simulated Annealing and the Genetic Algorithm are better suited to.

We next repeat these experiments for the Erd\H{o}s-R\'{e}nyi random graph shown in Figure~\ref{fig:testnetworks}~(b). The results are shown for bond dimensions $\chi=2$ and $\chi=10$ in Figure~\ref{fig:performancebyruntime}~(c),(d) respectively. Both Simulated Annealing and the Genetic Algorithm significantly outperform Greedy Search with a similar number of cost function evaluations. Moreover, they do so even with significantly fewer cost function evaluations. Intuitively, these non-local optimization methods are able to perform comparatively better with higher connectivity and less local structure.

Common to all of the panels of Figure~\ref{fig:performancebyruntime}, we see that Simulated Annealing and the Genetic Algorithm improve in bursts, separated by long plateaus. This makes it difficult to arrive at strong statements about the correct number of function evaluations to use with these algorithms, as there is no clear indication of whether the search will continue improving or not.

The large, order-of-magnitude nature of the bursts, however, suggests a heuristic to use in practice, which is that the search for better contraction sequences should be conducted for a time comparable to the run-time of the current best contraction sequence. We call this the ``time-remaining'' heuristic. In that way the search at most doubles the run-time if it yields no improvement, while still offering a chance of dramatic gains.

\begin{figure*}[t]
\centering
\includegraphics[width=\textwidth,page=5]{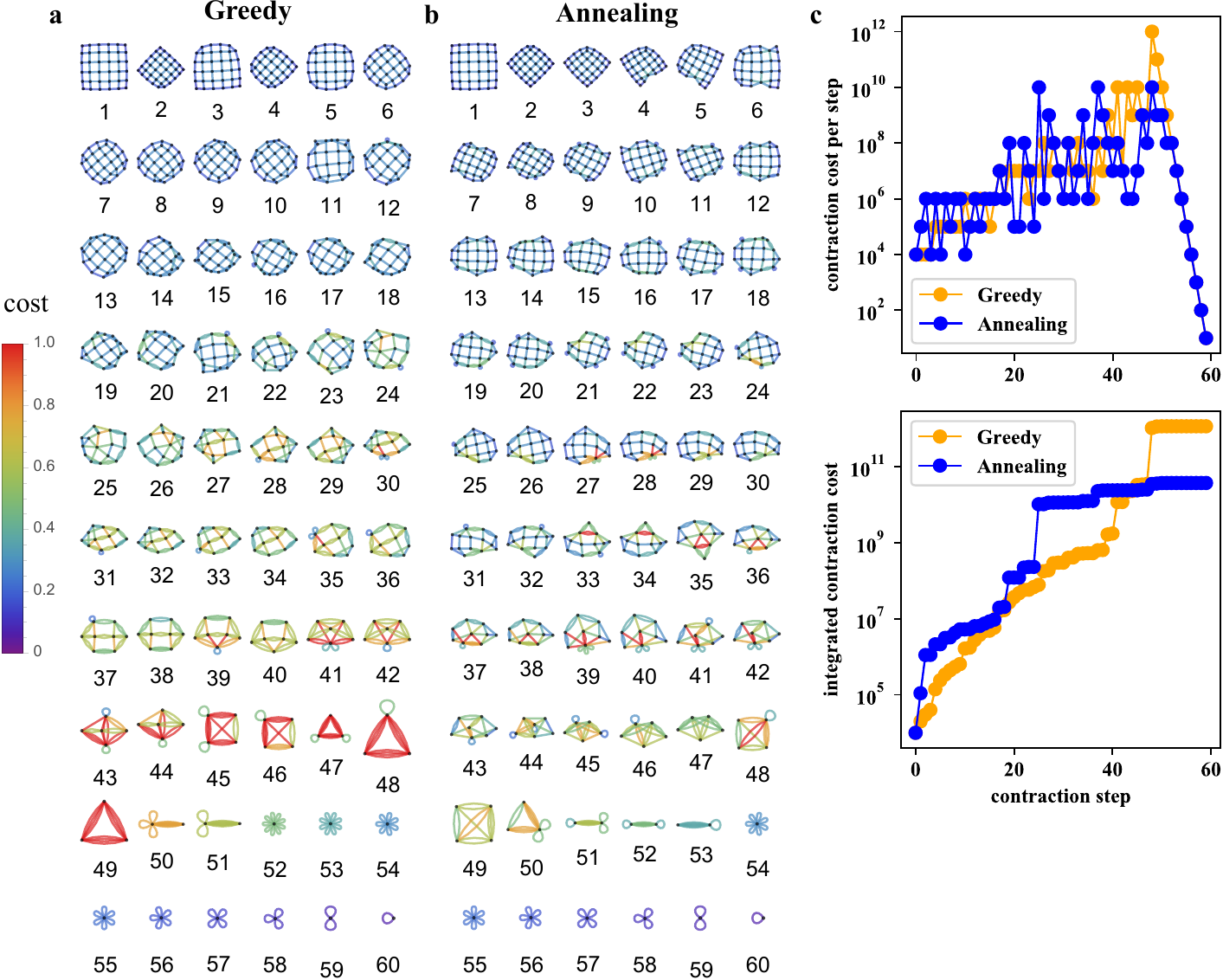}
\caption{Typical contraction sequences obtained for a square tensor network with $6^2$ nodes connected by edges with $\chi = 10$. We show the median sequence produced by 40 runs, where the median is taken with respect to the cost of the best sequence found in each run. The color of individual bonds indicates the cost of contracting them; we chose a color scale that is proportional to the fourth root of the contraction cost in order to enhance contrast. The color scale is the same for both algorithms. (a)~Greedy Search. (b)~Simulated Annealing. (c)~Comparison of the (accumulated) cost per contraction step.}
\label{fig:specificSequence1}
\end{figure*}

\begin{figure*}[t]
\centering
\includegraphics[width=\textwidth,page=6]{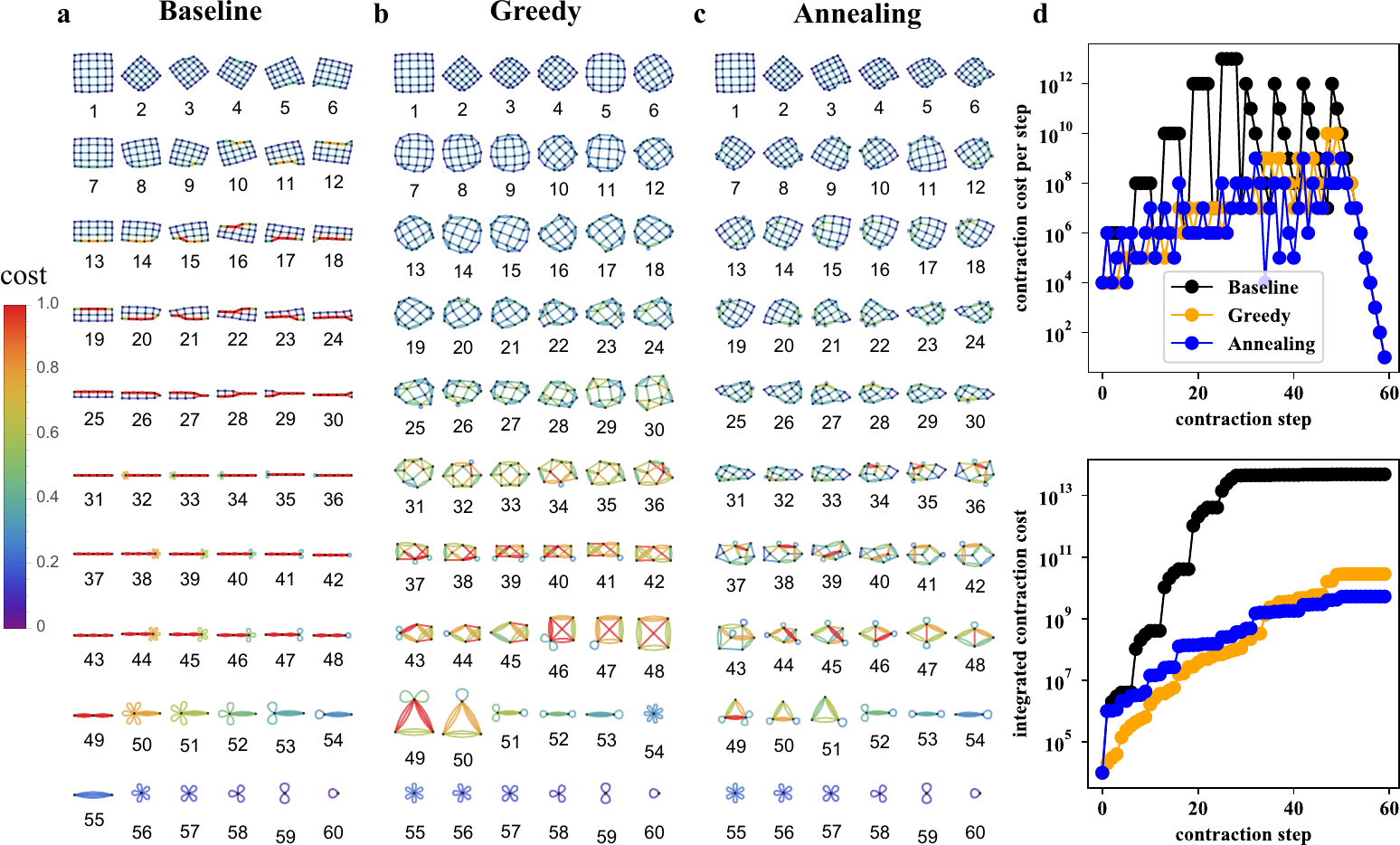}
\caption{Optimal contraction sequences obtained for a square tensor network with $6^2$ nodes connected by edges with $\chi = 10$. We show the minimum cost sequence produced by 40 runs. The color of individual bonds indicates the cost of contracting them; we chose a color scale that is proportional to the fourth root of the contraction cost in order to enhance contrast. The color scale is the same for all algorithms. (a)~Hand-crafted Sequence, (b)~Greedy Search. (c)~Simulated Annealing. (d)~Comparison of the (accumulated) cost per contraction step.}\label{fig:specificSequence2}
\end{figure*}

\subsection{Equal Run-Time}
For comparison purposes we do not adopt the time-remaining heuristic here. Rather we now fix the number of cost function evaluations used by each algorithm to be equal to that of the Greedy Search. This enables meaningful comparisons between algorithms with similar runtime contraints.

Figure~\ref{fig:performancesquare}~(a) shows the performance of Simulated Annealing, the Genetic Algorithm and Greedy Search on two-dimensional square tensor networks of varying sizes with bond dimension $\chi=2$. The left-most panel shows the median performance across $20$ runs of each non-deterministic algorithm, along with the global optimal result provided by Exhaustive Search. The middle-panel shows the relative standard deviation in performance for the same. Finally, the right-most panel shows the number of cost function evaluations used by the Exhaustive Search and Greedy Search. Simulated Annealing and the Genetic Algorithm were both allowed the same number of evaluations as Greedy Search.

We see relatively little variation in performance across these four algorithms, and to the point where we were able to use the Exhaustive Search the algorithms perform close to the global optimum. To the extent that there is a difference, it is for larger systems, for which Simulated Annealing significantly outperforms the other algorithms.

Interestingly, the relative standard deviation in cost is much larger with both Greedy Search and the Genetic Algorithm. We are not sure why the Genetic Algorithm has a high relative standard deviation. The high relative standard deviation of the Greedy Search is understandable, however: many choices are degenerate for the Greedy Search. Often a tensor network has many different contractions with the same immediate cost, and the same is true at higher search depths. Even though the immediate cost is degenerate, the long-term consequences of these choices on the network may be radically different, causing significant variance in total cost.

Figure~\ref{fig:performancesquare}~(b) shows the same experiment but with bond dimension $\chi=10$. We now see a larger spread between the algorithms. Here we have highlighted the so-called ``desktop limit'' of $\mathrm{cost} = 10^{16}$, which provides a rough bound on the cost of contractions that can reasonably be performed on a modern desktop computer limited to a day of runtime. (Note that if the cost is dominated by a single expensive contraction step the practical limit is somewhat lower, as tensors with $10^{16}$ elements are unlikely to fit into memory.) In particular, the gap between Greedy Search and Simulated Annealing is such that the former hits the desktop limit on systems roughly $10\%$ smaller than the latter, suggesting that with the improvements offered by Simulated Annealing it should be possible to contract larger tensor networks than were previously possible.

The larger bond dimension also brings about an increased relative standard deviation in their performance. The increased relative standard deviation comes about because the cost is now more sensitive to the few most expensive contraction steps, and so becomes more sensitive to the (discrete) ranks of tensors as the contraction proceeds.

We next repeated these experiments with Erd\H{o}s-R\'{e}nyi random graphs of varying size. The results are shown in Figure~\ref{fig:performancerandom}. For small systems the algorithms all find nearly-optimal contraction sequences. As the system size increases above $10-11$ nodes a large difference emerges which grows until the Greedy Search performs a factor of $10-100$ worse than Simulated Annealing, which in turn performs a factor of $10$ or so worse than optimal. The relative standard deviation in performance across runs is generally larger than in the square network cases, particularly for Simulated Annealing. The overall increase can be understood as being due to increased variance in tensor ranks making the cost more sensitive to the precise contraction sequence. We are not sure why this affects Simulated Annealing more than the other algorithms, though it may indicate that with random graphs the dual annealing implementation is less able to exploit the structure of the network in its local search steps.

\begin{figure*}[t]
\centering
\includegraphics[width=\textwidth,page=7]{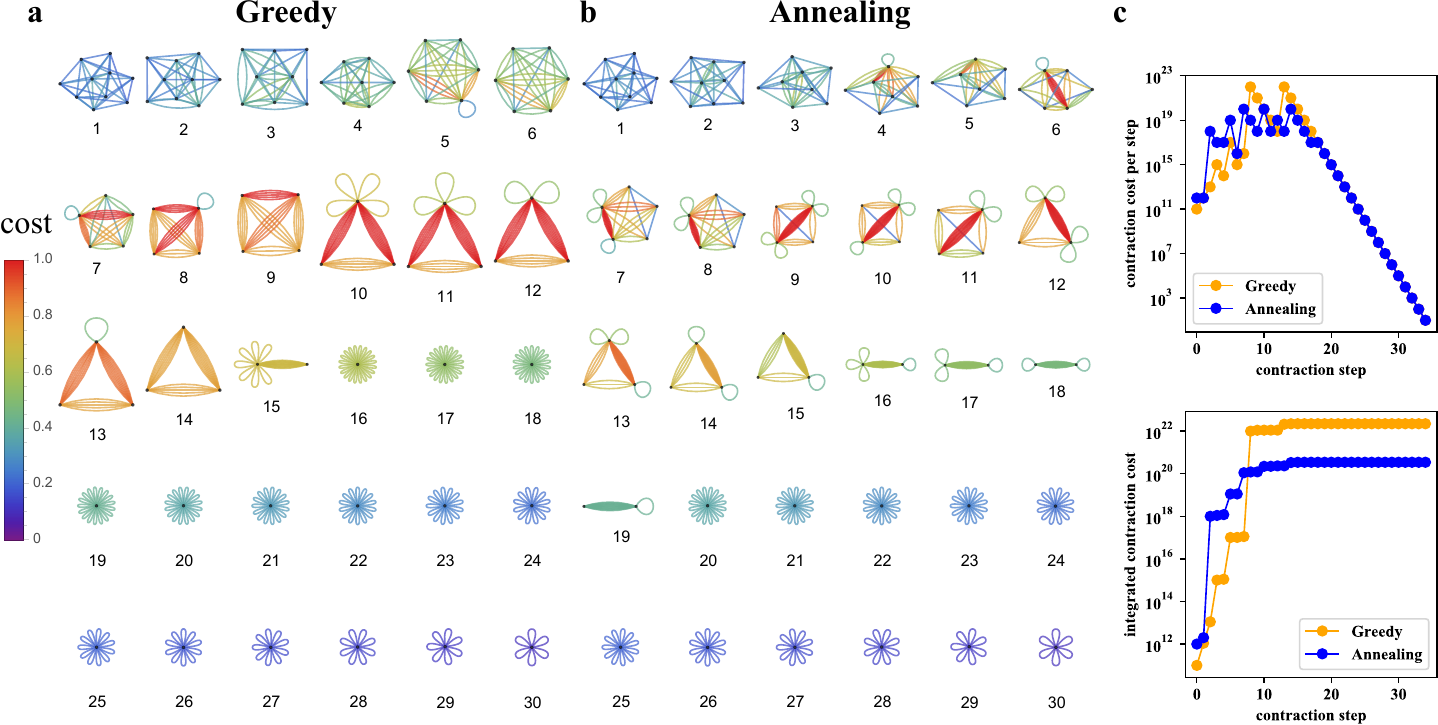}
\caption{Optimal contraction sequences obtained for a random tensor network with $10$ nodes connected by edges with $\chi = 10$ (and a connectivity of $80\%$). We show the minimum cost sequence produced by 40 runs. The color of individual bonds indicates the cost of contracting them, we chose a color scale that is proportional to the eighth root of the contraction cost in order to enhance contrast. The color scale is the same for both algorithms. (a)~Greedy Search. (b)~Simulated Annealing. (c)~Comparison of the (accumulated) cost per contraction step.}\label{fig:specificSequence2}
\label{fig:optrandom}
\end{figure*}

\section{Contraction Sequences}
\label{sec:seq}
To understand how Simulated Annealing comes to outperform the Greedy Search it is useful to examine a typical contraction sequence produced by each algorithm. In Figure~\ref{fig:specificSequence1} we show the median best contraction sequence of each algorithm taken across 40 runs for a square network with linear size $L=6$ and bond dimension $\chi=10$. We depict the contraction sequence by showing the tensor graph before each contraction step. The colors of individual bonds are proportional to the fourth root of the contraction cost, with red indicating higher cost and blue indicating lower cost.

The Simulated Annealing sequence is roughly $100$ times less expensive than that of the Greedy Search. As expected for this large bond dimension, both algorithms have costs which are dominated by three or fewer steps [Fig~\ref{fig:specificSequence1}~(c)]. From Figure~\ref{fig:specificSequence1}~(a)-(b) we see that Simulated Annealing always leaves itself a comparatively low-cost option, while Greedy Search exhausts all such options and is forced into costly contraction steps.

Arranging to retain lower-cost options is inherently a global optimization process, because the cost of contracting an edge is strongly dependent on the stage at which it is contracted and on the contraction order leading up to that point. This explains why Simulated Annealing is able to achieve this task while Greedy Search is not: the early costlier contractions that Simulated Annealing performs act to reduce the cost of the most expensive steps towards the end.

A hint is provided by the Simulated Annealing sequence between steps 26 and 29, and again between steps 32 and 43. In both cases, one or more bonds emerge which involve expensive contractions. In the first instance Simulated Annealing performs the expensive contraction, which enables lower-cost options afterwards. In the second instance it merges nearby nodes into those adjacent to the expensive bond (e.g. 35 - 36). In doing so it produces self-loops on the adjacent nodes which, upon elimination, reduce the cost of the deferred expensive step.

We next turn to the best contraction sequence produced by each algorithm. In Figure~\ref{fig:specificSequence2}~(a)-(c) we show the best contraction sequence of each algorithm taken across 40 runs for the same network as in Figure~\ref{fig:specificSequence1}. We also show for comparison a typical hand-crafted contraction sequence similar to the corner transfer matrix method commonly used with PEPS. In this sequence rows are contracted together repeatedly until just one remains, at which point that row is contracted down to a point. We see that both Greedy Search and Simulated Annealing significantly outperform the hand-crafted sequence. Both algorithms manage this by producing fewer high-rank tensors, which is enabled by ``contracting inwards'' from the perimeter rather than working with a single edge the whole time. Doing so reduces the number of extra bonds accumulated by each tensor on the edge, which holds the rank down.

In Figure~\ref{fig:specificSequence2}~(d) we see that, like in the median case, there are just a few steps which together dominate the cost of the best contraction sequence for each algorithm. However, unlike in the median case, in the best case the costs of the different contraction sequences are of the same order of magnitude, which is a factor of $10$ or so better than the median case for the Simulated Annealing algorithm. This suggests that the best case for the Greedy Search is a bigger improvement over the median case than the best case for the Simulated Annealing algorithm is over its median case.

We can understand this improvement by noting that in the early stages of the Greedy contraction sequence there is significant degeneracy between the various least-expensive contractions. This results in a wide variety of different possible states following the first 10 or so contraction steps. The typical such state is evidently much harder to continue contracting than the best such state. This conclusion highlights the importance of optimizing globally over the whole contraction sequence, and not just locally as Greedy Search does.

Finally, in Figure~\ref{fig:optrandom} we show the best contraction sequence of Greedy Search and Simulated Annealing across 40 runs for an Erd\H{o}s-R\'{e}nyi random graph. We see again that Simulated Annealing does a much better job of preserving comparatively good options throughout the contraction, while Greedy Search exhausts its cheap contraction options early on and is forced into a run of very expensive contraction steps. Unfortunately the random structure of this graph exacerbates our earlier challenge interpreting the Simulated Annealing contraction sequence, and it is not clear exactly what choices it is making that enable such good long-run performance. Nevertheless, the performance is remarkable: Simulated Annealing finds a contraction sequence that is $100$ times faster than that of Greedy Search, and does so with the same number of cost function evaluations.

\section{Conclusions}
\label{sec:conc}
We have optimized tensor network contraction sequences using four different algorithms, namely Exhaustive Search, Greedy Search, Simulated Annealing, and a Genetic Algorithm. The first two of these are commonly used in contracting tensor networks, while to our knowledge the latter two have not been used in this domain. We find that Simulated Annealing significantly outperforms Greedy Search, both in the best case and on average. In many cases the cost of the contraction sequence found by Simulated Annealing is orders of magnitude lower than that of Greedy Search with a comparable amount of search time. This advantage grows larger with network size, and is most notable on networks with structure such as the square lattice.

With additional search time Simulated Annealing performs even better, often by a large enough margin to justify the extra time spent optimizing the contraction sequence. This suggests a potential strategy to use in practice, our ``time-remaining'' heuristic, in which one optimizes the contraction sequence until the time spent optimizing is comparable to the cost of the current best known contraction sequence.

While contracting large tensor networks generally requires approximation, our results suggest that exact contraction may be viable for larger networks than previously thought. This may provide benefits even for approximate contraction methods such as Tensor Network Renormalization, which often rely on repeatedly and exactly contracting moderate-sized networks to produce inputs into the approximation scheme~\cite{PhysRevLett.115.180405}.

Unfortunately we have been unable to extract any further intuition from these contraction sequences. They do not appear to lead to design principles we may use to craft custom sequences for particular classes of networks. In practice, however, this may not matter: algorithmically-generated contraction sequences may well suffice, particularly if they are more efficient than hand-crafted or heuristically-guided ones.

\section*{Acknowledgements}
We thank Miles Stoudenmire for early conversations relating to this work. F.~S. thanks Noa Nabeshima for suggesting the use of genetic algorithms and helpful discussions. F. S. acknowledges support from the Swiss National Science Foundation (grant number: 200021\_169061). This research was supported in part by the National Science Foundation under Grant No. NSF PHY-1748958, by the Gordon and Betty Moore Foundation through Grant GBMF7392, by the Heising-Simons Foundation, and by the Flatiron Institute of the Simons Foundation.

\bibliography{tensor_refs}

\end{document}